\documentclass{article} 
\usepackage[utf8]{inputenc}
\usepackage{geometry}
\usepackage{graphicx}
\usepackage{booktabs}
\usepackage{array}
\usepackage{needspace}
\usepackage{placeins}
\usepackage{amsmath, amssymb}
\usepackage{authblk} 
\usepackage{arydshln}
\usepackage[numbers,sort&compress]{natbib}
\usepackage[colorinlistoftodos]{todonotes}
\usepackage[colorlinks=true,
            linkcolor=cyan,
            citecolor=cyan,
            urlcolor=cyan]{hyperref}

\usepackage{xcolor}
\definecolor{RoyalBlue}{RGB}{0, 50, 100}   
\definecolor{SoftGray}{RGB}{240, 240, 240} 
\definecolor{LinkBlue}{RGB}{0, 80, 160}  
\definecolor{blueviolet}{RGB}{138,43,226}

\usepackage[most]{tcolorbox}

\newtcolorbox{abstractbox}{
  colback=SoftGray,        
  colframe=SoftGray,       
  arc=3mm,                 
  boxrule=0mm,             
  left=4mm, right=4mm, top=4mm, bottom=4mm, 
  fontupper=\sffamily\small, 
  before skip=1em, after skip=2em
}

\newtcolorbox{insightblock}{
  colback=blueviolet!5,   
  colframe=blueviolet!50!black!50!,    
  boxrule=0.5mm,       
  arc=2mm,            
  left=0pt,           
  right=8pt,           
  top=8pt,          
  bottom=8pt,        
}

\usepackage{graphicx}
\graphicspath{{./Figures/}}

\title{Learning with Limited Antigenic Labels: Semi-supervised Prediction of Influenza A Antigenic Relationships with Protein Language Models}

\author[1]{Yanhua Xu\thanks{Correspondence: alysiayx@outlook.com}}
\affil[1]{\small Department of Computer Science, University of Liverpool, UK}
\date{}

\begin{document}

\maketitle

\begin{abstractbox}

\begin{center}
\textbf{\normalsize Abstract}
\end{center}

\textbf{Background}
Influenza A viruses (IAVs) undergo frequent antigenic changes that require regular vaccine updates. Because haemagglutination inhibition (HI) assays are labour-intensive, antigenic measurements are available for far fewer samples than sequence data. Here, we combined semi-supervised learning (SSL) with pre-trained protein language models (PLMs) to predict antigenic relationships when labelled data are limited.

\vspace{0.5em}

\textbf{Methods}
ESM-2, ProtVec, ProtT5, and ProtBert encoded sequences from the HA1 domain of haemagglutinin (HA) in H1N1, H3N2, H5N1, and H9N2. We compared self-training and label spreading with supervised random forest (RF) and support vector machine (SVM) models using nested cross-validation with 25\%, 50\%, 75\%, or 100\% of the training labels retained.

\vspace{0.5em}

\textbf{Results}
Self-training slightly improved RF performance in most settings but reduced SVM performance in nearly all settings. At 25\% label availability, self-training raised mean macro-F\textsubscript{1} from 0.791 to 0.798 for ProtVec-RF and from 0.776 to 0.778 for ESM-2 RF. Label spreading classified antigenic relationships less effectively than supervised RF across all protein representations and training-label proportions (mean macro-F\textsubscript{1}, 0.768--0.832). For RF models using ESM-2 or ProtBert, additional training labels improved classification of H3N2 antigenic relationships, while H9N2 remained the most difficult subtype to classify.

\vspace{0.5em}

\textbf{Conclusion}
Self-training offered small improvements over supervised RF but generally reduced SVM performance. RF models struggled most with H9N2, where antigenically similar pairs were often misclassified as variants.

\vspace{0.5em}


\textbf{Keywords}
Machine Learning; Semi-supervised Learning; Influenza A viruses; Antigenicity; Protein Language Models

\end{abstractbox}
%

\section{Introduction}

Influenza A viruses (IAVs) undergo frequent antigenic changes as mutations accumulate in their surface proteins, haemagglutinin (HA) and neuraminidase (NA). These changes can alter recognition by antibodies acquired through infection or vaccination, creating a continuing need to monitor circulating viruses and update vaccine composition. Surveillance combines sequence analysis with experimental measurements of antigenic relationships. Haemagglutination inhibition (HI) assays help assess the match between vaccine and circulating strains \citep{ndifon2009use, gupta2006quantifying}, but remain labour-intensive, require specific animal antisera, and can vary between laboratories. Sequencing has become faster and less expensive, leaving many virus pairs with available sequences but no measured antigenic relationship.

Sequence-based prediction addresses this gap by estimating antigenic relationships from viral sequences. Earlier studies classified pairs as antigenically Similar or Variant using models trained on HI-based class assignments \citep{liao2008bioinformatics, yin2018predicting}. Those assignments provide the labelled examples for supervised learning, but the available sequences also describe many pairs without an experimental label. Our dataset contains 2,170 labelled pairs and 15,283 unlabelled pairs. Semi-supervised learning (SSL) brings these additional pairs into training, allowing the classifier to use sequence variation beyond the experimentally characterised relationships.

What the classifier can learn from either group depends on how the sequences are represented. Earlier influenza studies used residue-level, regional, and structural descriptors to describe differences between viruses \citep{yin2018predicting, peng2017universal, qiu2022univ}. Pre-trained protein models offer representations learned from much broader sequence collections. ProtVec assigns vectors to short amino acid segments, while protein language models (PLMs), including ESM-2, ProtBert, and ProtT5, represent amino acids in their surrounding sequence context \citep{asgari2015continuous, lin2023evolutionary, elnaggar2021prottrans}. The antigenic classifier can then use these embeddings together with HI-based labels to distinguish Similar from Variant pairs.

We combine these representations with self-training and label spreading, which use unlabelled pairs in different ways. Self-training reuses selected model predictions as labels, while label spreading shares class information between samples with similar representations \citep{zhou2003learning}. The key questions are whether these methods improve on supervised learning with the same number of labels and how their scores compare with supervised learning using all available labels.

These comparisons span subtypes with different numbers of labelled pairs and different proportions of Similar and Variant relationships. High overall scores can therefore accompany poorer predictions in a smaller subtype or in one of the two classes. We use macro-F\textsubscript{1} to give both classes equal weight and report the subtype results separately. For H9N2, the class-specific scores show whether errors arise mainly from failure to recognise antigenic similarity or antigenic variation.

Here, we evaluate ESM-2, ProtVec, ProtT5, and ProtBert representations using retrospective data from H1N1, H3N2, H5N1, and H9N2. Supervised classifiers are compared with self-training and label spreading at 25\%, 50\%, 75\%, and 100\% training-label availability, using nested cross-validation to separate model tuning from test evaluation. Alongside overall performance across all models, we compare individual subtypes using supervised and self-training random forest (RF) with ESM-2 or ProtBert, then examine the two antigenic classes within H9N2.

\section{Related Work}

Sequence-based antigenic prediction combines HA sequence features with statistical and machine learning models. Liao et al.\ \citep{liao2008bioinformatics} used scoring and regression methods to predict antigenic variants of A/H3N2, while Degoot et al.\ \citep{degoot2019predicting} related sequence and antigenic data through a model based on biophysical ideas. Yin et al.\ \citep{yin2018predicting} combined residue-level, regional, and epitope features in a stacking framework for H1N1 viruses from epidemic and pandemic periods. Their ensemble improved classification over individual predictors, with performance also differing between the historical periods represented in the data.

Other studies have explored sequence encodings together with neural classifiers. Zhou et al.\ \citep{zhou2018context} developed a context-free encoding scheme for diverse influenza A viruses; IAV-CNN \citep{yin2021iav} used a two-dimensional convolutional neural network to classify antigenic variants. For A/H3N2, Xia et al.\ \citep{xia2021deep} combined CNN and bidirectional recurrent models. These studies used different combinations of sequence representations and classifiers. We compare the protein representations using the same classifiers, including when only a fraction of the training labels is available.

The studies also differ in the viruses they cover. PREDAC-H1pdm \citep{liu2023development} predicted antigenic relationships and identified clusters in post-2009 pandemic H1N1 viruses, using those clusters to describe variation over time. PREDAV-FluA \citep{peng2017universal} addressed broader HA subtype coverage with regional band-based features, while Univ-Flu \citep{qiu2022univ} combined three-dimensional structure-based descriptors with random forest classifiers.

Beyond this difference in coverage, studies vary in what they predict. Joint random forests have been used to estimate continuous antigenic distances from HA sequences \citep{yao2017predicting}. Wang et al.\ \citep{wang2018predicting} filled gaps in antigenic measurements through matrix completion based on antigen and antiserum similarity, while Peng et al.\ \citep{peng2023prediction} predicted H3N2 antigenic distances with attribute network embeddings. These methods estimate the antigenic distance between two viruses. Our study predicts whether the two viruses are antigenically similar or different.

Pre-trained protein models provide sequence features learned from large protein collections. Asgari and Mofrad \citep{asgari2015continuous} introduced ProtVec, which assigns distributed vectors to short amino acid segments. Later Transformer models, including ESM-2 \citep{lin2023evolutionary} and the ProtTrans models ProtBert and ProtT5 \citep{elnaggar2021prottrans}, made those representations depend on sequence context. ProtTrans used embeddings learned from unlabelled proteins in tasks such as secondary-structure and subcellular-localisation prediction. These models therefore offer representations that differ in their sequence units, use of context, and pre-training collections.

The use of unlabelled data also extends to biological tasks with limited experimental annotation. Wu et al.\ \citep{liu2024mmaple} developed MMAPLE, a semi-supervised meta-learning framework for understudied molecule--protein interactions. In influenza antibody research, Wang et al.\ \citep{wang2024explainable} curated HA antibodies and developed a memory B cell language model (mBLM) for antibody specificity prediction. The present study focuses on antigenic relationships between virus pairs, combining pre-trained sequence representations with SSL to use pairs that lack an experimental class assignment.

\section{Materials and Methods}

\subsection{Data Set}
We compiled antigenic data from public sources and published studies \citep{degoot2019predicting, zhou2018context, yin2021iav, liao2008bioinformatics, yin2018predicting, peng2017universal}. The dataset comprises 2,170 labelled virus pairs involving 329 unique sequences from the HA1 domain of HA across subtypes H1N1, H3N2, H5N1, and H9N2.

Antigenic similarity was quantified using the Archetti-Horsfall distance ($d_{DV}$) \citep{archetti1950persistent, ndifon2009use, gupta2006quantifying, lee2004predicting}, derived from HI titres:
\begin{equation}
d_{DV} = \sqrt{\frac{H_{DD} \times H_{VV}}{H_{DV} \times H_{VD}}}
\end{equation}
where $H_{XY}$ represents the HI titre between virus $X$ and antiserum raised against virus $Y$.

A common cutoff of 4 was used across all four subtypes. Pairs with $d_{DV}<4$ were classified as antigenically \textit{Similar}, and those with $d_{DV}\geq4$ as \textit{Variant} \citep{liao2008bioinformatics}. A further 15,283 within-subtype sequence pairs without direct HI measurements were retained as unlabelled data for SSL. Table~\ref{tab_data} lists the sample counts for each subtype.

\begin{table}[ht]
\centering
\caption{Dataset composition by subtype. Total pairs include labelled Similar and Variant pairs and unlabelled pairs.}
\label{tab_data}
\begin{tabular}{@{}cccccc@{}}
\toprule
\textbf{Subtypes}& \textbf{\# sequences} & \textbf{Total pairs} & \textbf{\# similar} & \textbf{\# variant}  & \textbf{\# unlabelled}\\ \midrule
H1N1              &                    152&              11,448&                   483 &                   851 &10,114\\
H3N2              &                    61&               1,826&                  125 &                   142 &1,559\\
H5N1              &                    87&               3,687&                  186 &                   265 &3,236\\
H9N2              &                     29&              400&                  31 &                     87 &282 \\
Total                &                    329&               17,453&                  825&                  1,345&15,283\\ \bottomrule
\end{tabular}
\end{table}

\subsection{Protein Embeddings}
\label{sec_protein_embeddings}

We encoded HA1 sequences with ESM-2, ProtBert, ProtT5, and ProtVec. The three Transformer models produce amino acid representations that depend on surrounding sequence context, whereas ProtVec uses Word2Vec-based vectors for short amino acid segments.

\textbf{ESM-2.}
For ESM-2, we used the \texttt{facebook/esm2\_t33\_650M\_UR50D} checkpoint, a 33-layer Transformer with 650 million parameters and 1280-dimensional embeddings. ESM-2 was pre-trained on UniRef50 using masked language modelling \citep{lin2023evolutionary}.

\textbf{ProtBert.}
ProtBert is a BERT-based model in the ProtTrans family, pre-trained on UniRef100 using masked language modelling \citep{elnaggar2021prottrans}. It produces a 1024-dimensional embedding for each amino acid using sequence context in both directions.

\textbf{ProtT5.}
We obtained 1024-dimensional encoder representations in half-precision from ProtT5-XL-U50, a T5-based model trained on UniRef50 \citep{elnaggar2021prottrans}. T5 uses an encoder--decoder architecture and learns to reconstruct sequences during pre-training.

\textbf{ProtVec.}
ProtVec represents amino acid 3-grams using a Word2Vec-inspired model trained on Swiss-Prot sequences \citep{asgari2015continuous}. Each tripeptide has the same 100-dimensional vector wherever it occurs. ProtVec served as a lower-dimensional baseline for the contextual Transformer embeddings.

All sequences were padded or truncated to uniform lengths. For the Transformer models, mean-pooling of the final hidden-layer representations produced fixed-size vectors for classification. Table~\ref{tab_pretrained} summarises the embedding dimensions, tokenisation, pre-training datasets, and objectives.

\begin{table}[ht]
    \centering
    \caption{Architecture and pre-training specifications of the protein representations.}
    \label{tab_pretrained}
    \small
    \setlength{\tabcolsep}{4pt}
    \renewcommand{\arraystretch}{1.2}
    \begin{tabular*}{\linewidth}{@{\extracolsep{\fill}}
        >{\raggedright\arraybackslash}p{1.6cm}
        >{\raggedright\arraybackslash}p{0.9cm}
        >{\raggedright\arraybackslash}p{1.55cm}
        >{\raggedright\arraybackslash}p{1.9cm}
        >{\raggedright\arraybackslash}p{2.2cm}
        >{\raggedright\arraybackslash}p{2.5cm}
        >{\raggedright\arraybackslash}p{1.7cm}@{}}
    \toprule
    \textbf{Model} & \textbf{Dim.} & \textbf{Sequence unit} &
    \textbf{Training data} & \textbf{Tokenisation} &
    \textbf{Masking} & \textbf{Objective} \\
    \midrule
    \mbox{ESM2-33} & 1280 & Amino acid & UniRef50 (65M) &
    Standard AA-level & 15\% MLM (BERT-style) & Masked LM \\
    \addlinespace[5pt]
    ProtBert & 1024 & Amino acid & UniRef100 (217M) &
    AA-level (space-delimited) & 15\% MLM (80-10-10 split) & Masked LM \\
    \addlinespace[5pt]
    ProtT5 & 1024 & Amino acid & UniRef50 (45M) &
    AA-level (space-delimited) & Span corruption (15\%) & Seq2Seq denoising \\
    \addlinespace[5pt]
    ProtVec & 100 & Tripeptide & Swiss-Prot (549K) &
    3-mer segmentation & N/A & Skip-gram \\
    \bottomrule
    \end{tabular*}
    \par\vspace{4pt}
    \begin{minipage}{\linewidth}
    \footnotesize Dim., embedding dimension; AA, amino acid; MLM, masked language modelling; LM, language modelling.
    \end{minipage}
\end{table}
\begin{samepage}
\subsection{Semi-supervised Learning Algorithms} 

We compared self-training with RF or support vector machine (SVM) classifiers and graph-based label spreading. Supervised RF and SVM provided the corresponding baselines for self-training.

\end{samepage}

\subsubsection{Self-training}
Self-training starts from a classifier fitted to labelled pairs, then reuses selected predictions for unlabelled pairs as pseudo-labels. Under threshold-based selection, examples with prediction confidence above $\tau$ enter the labelled training set, and the classifier is refitted as this set expands. Table~\ref{tab_tunning_s3} gives the candidate selection criteria and iteration settings.

\subsubsection{Label Spreading}
Label spreading connects labelled and unlabelled samples in a graph according to similarities between their protein representations \citep{zhou2003learning}. Class information from the observed labels propagates along those connections to assign classes to unlabelled samples. The hyperparameter search settings are given in Table~\ref{tab_tunning_s3}.

\subsection{Implementation}
The supervised models comprised RF, SVM, convolutional neural network (CNN), and bidirectional gated recurrent unit (BiGRU) classifiers. Together with self-training RF, self-training SVM, and label spreading, these formed the seven models in the comparison. Each was evaluated with the four protein representations at four training-label proportions, giving 112 settings.

\begin{figure}[!htbp]
    \centering
\includegraphics[width=0.5\linewidth]{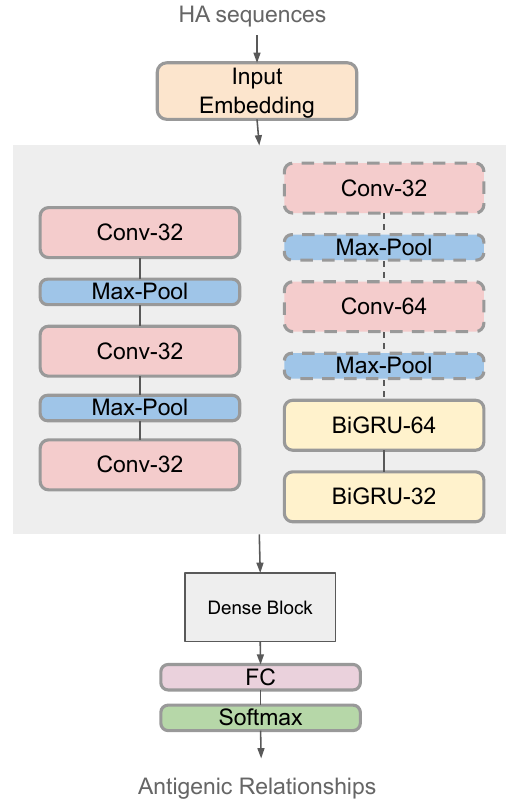}
    \caption{CNN and BiGRU architectures for classifying HA1 sequence embeddings.}
    \label{fig_cnn_bigru_architecture}
\end{figure}

All models used the embeddings described in Section~\ref{sec_protein_embeddings}. The neural classifiers were implemented in TensorFlow/Keras and the conventional classifiers and SSL methods in scikit-learn. The CNN and BiGRU architectures are shown in Figure~\ref{fig_cnn_bigru_architecture}, with the hyperparameter search spaces in Table~\ref{tab_tunning_s3}.

\begin{table}[!htbp]
\caption{Hyperparameter search spaces.}
\label{tab_tunning_s3} 
\centering
\footnotesize
\renewcommand{\arraystretch}{1.2}
\begin{tabular*}{\linewidth}{l@{\extracolsep{\fill}}@{}ll@{}}
\toprule
\multicolumn{1}{c}{\textbf{ Algorithms }} & \multicolumn{1}{c}{\textbf{Hyperparameters}} \\
 \midrule

SVM &
\begin{tabular}[c]{@{}l@{}}
$C$ = [$10^{-6}$, $10^{1}$] \\ 
$\gamma$ = [$10^{-6}$, $10^{1}$] \end{tabular} \\

\hdashline

RF & 
\begin{tabular}[c]{@{}l@{}}
$n_\text{estimators}$ = $10, 50, 100, 150, 200$\\ 
$\text{max\_depth} \in \{5, 10, 15, 20, \text{None}\}$
\end{tabular} \\

\hdashline

CNN & 
\begin{tabular}[c]{@{}l@{}}
$\text{num\_filters} \in \{32, 64, 128, 512\}$ \\
$\text{learning\_rate} \in \{10^{-5}, 10^{-4}, 10^{-3}, 10^{-2}\}$ \\
$\text{batch\_size} \in \{32, 64, 128\}$ \\
$\text{epochs} = 500$ \\
$\text{kernel\_size} \in \{3, 5\}$
\end{tabular} \\

\hdashline

BiGRU & 
\begin{tabular}[c]{@{}l@{}}
$\text{units} \in \{32, 64, 128, 256, 512\}$ \\
$\text{learning\_rate} \in \{10^{-5}, 10^{-4}, 10^{-3}, 10^{-2}\}$ \\
$\text{batch\_size} \in \{32, 64, 128\}$ \\
$\text{epochs} = 500$
\end{tabular} \\

\hdashline

Label Spreading & 
\begin{tabular}[c]{@{}l@{}}
kernel = knn \\
$\alpha \in \{0.1, 0.2, 0.3\}$ \\
$n_\text{neighbors} \in \{3, 5, 7, 11, 20, 30, 40, 50, 75, 100\}$ \\
$\text{max\_iter} \in \{20, 25, 30, 35, 40, 45, 50\}$
\end{tabular} \\

\hdashline

Self-training & 
\begin{tabular}[c]{@{}l@{}}
$\text{threshold} \in \{0.50, 0.55, \dotsc, 0.95, 0.99\}$ \\
criterion = \{threshold, k-best\} \\
$k_\text{best} \in \{3, 5, 10, 15\}$ \\
$\text{max\_iter} \in \{5, 10, 15, 20\}$
\end{tabular} \\

\bottomrule

\end{tabular*}
\end{table}

\Needspace{11\baselineskip}
\subsection{Model Evaluation}

\begin{figure*}[ht]
\centering
\includegraphics[width=\linewidth]{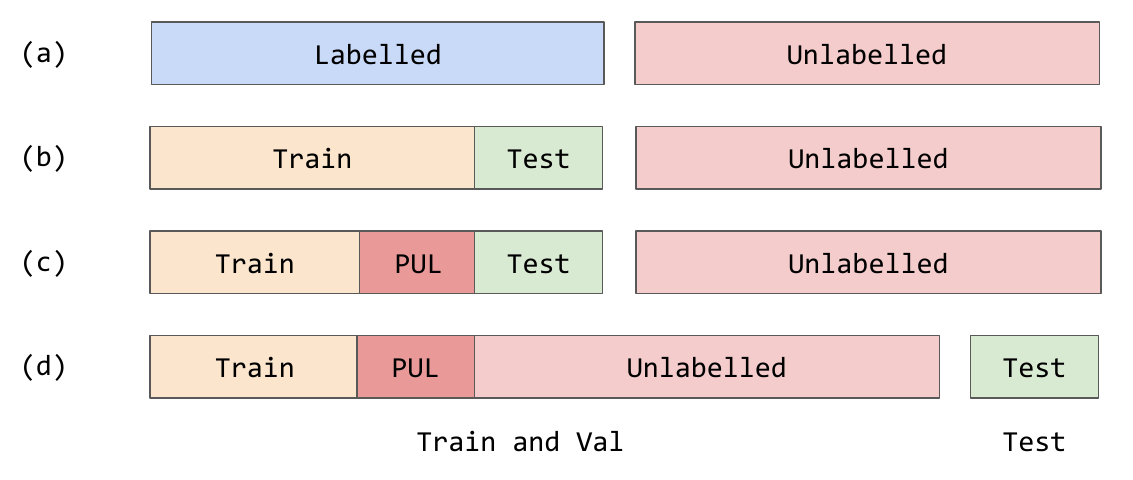}
\caption{Data composition during cross-validation; the inner hyperparameter-tuning loop is omitted.
(a) Full dataset composition. 
(b) Labelled samples are split into training and test sets. 
(c) Some training labels are masked to create pseudo-unlabelled (PUL) examples.
(d) SSL training data combine labelled, pseudo-unlabelled, and originally unlabelled examples.}
\label{fig_ssl_eval_process}
\end{figure*}

We evaluated the models using nested cross-validation, with five outer folds for testing and four inner folds for hyperparameter selection. Within each training set, we retained 25\%, 50\%, 75\%, or 100\% of the observed labels and treated pairs with masked labels as pseudo-unlabelled data for SSL. The originally unlabelled pairs were included in SSL training at every label proportion, including when all observed training labels were retained. Test labels remained unchanged and were used only for evaluation (Figure~\ref{fig_ssl_eval_process}).

We compared SSL with supervised learning at the same label proportion and representation, keeping the base classifier fixed for self-training. We also compared self-training RF using fewer labelled pairs with supervised RF trained on all available training labels. Overall scores cover all models. For supervised and self-training RF with ESM-2 or ProtBert, we also report results separately for each subtype. Within H9N2, we further report F\textsubscript{1} for each antigenic class. The evaluation metrics are defined in Section~\ref{sec_evaluation_metric}.

\Needspace{14\baselineskip}
\subsection{Evaluation Metric}
\label{sec_evaluation_metric}

Performance was evaluated with macro-F\textsubscript{1}, the mean of the F\textsubscript{1} scores for the Similar and Variant classes. The class-specific F\textsubscript{1} score is defined as:

\begin{equation}
F_1 = \frac{2 \cdot \mathrm{Precision} \cdot \mathrm{Recall}}{\mathrm{Precision} + \mathrm{Recall}} = \frac{2 \cdot \mathrm{TP}}{2 \cdot \mathrm{TP} + \mathrm{FP} + \mathrm{FN}}
\end{equation}

We report macro-F\textsubscript{1} overall and separately for each subtype. For H9N2, we further report the F\textsubscript{1} score for each antigenic class. Results for each protein representation also include averages across all models and within the supervised and SSL groups.

\section{Results}

\subsection{Performance under Limited Labels}
\begin{figure*}[!htb]
\centering
\includegraphics[width=\linewidth]{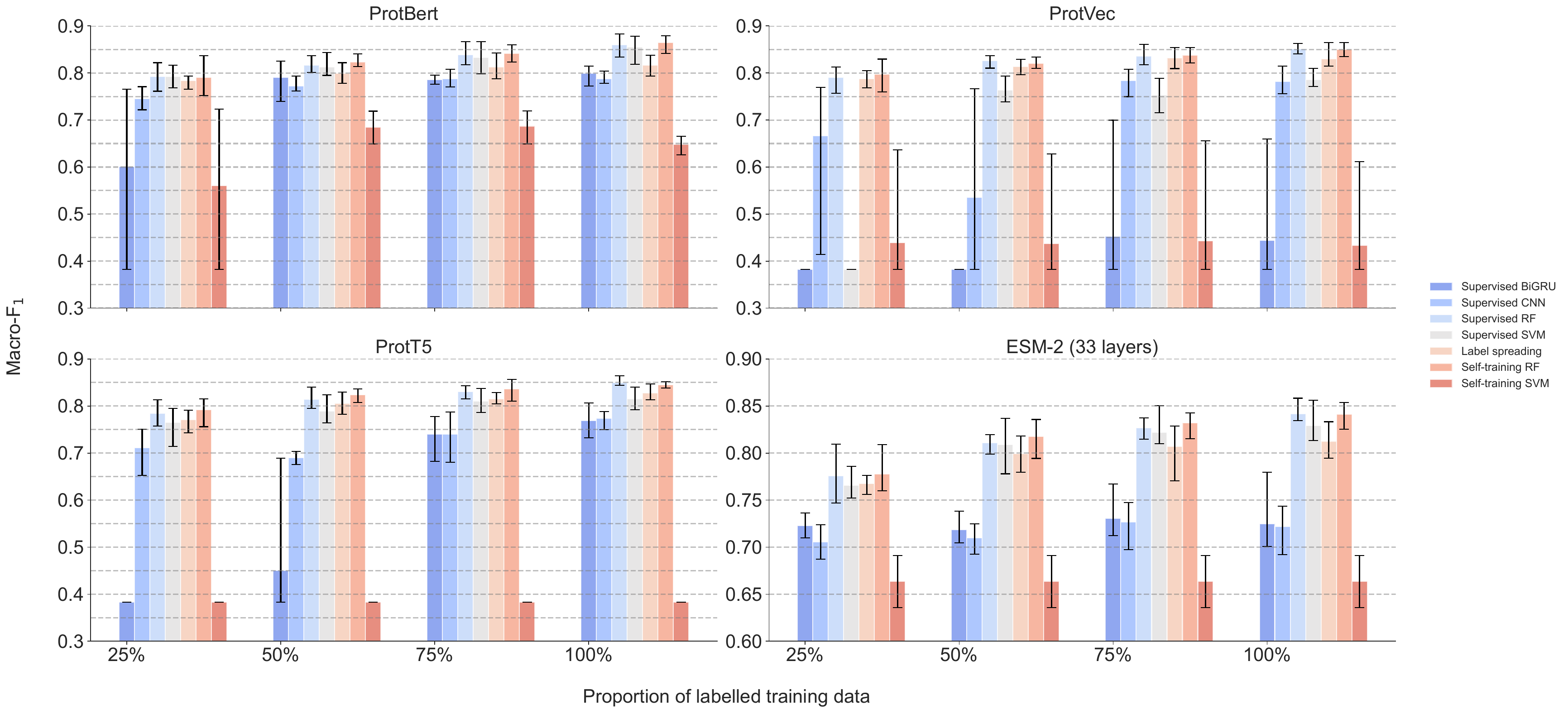}
\caption{Performance of supervised and semi-supervised models with 25\%, 50\%, 75\%, or 100\% of training labels retained. Bars show mean macro-F\textsubscript{1} across the five outer folds; error bars show the 2.5th--97.5th percentiles of the five fold scores.}
\label{fig_results_all_groups}
\end{figure*}

RF generally performed slightly better with self-training than with supervised learning. The largest increase in mean macro-F\textsubscript{1} was 0.009, and the largest decrease was 0.006 (Figure~\ref{fig_results_all_groups}). At 25\% label availability, self-training raised mean macro-F\textsubscript{1} from 0.791 to 0.798 with ProtVec but lowered it from 0.793 to 0.790 with ProtBert. With 75\% of training labels retained, self-training RF scored higher than supervised RF with all four protein representations. When all training labels were retained, self-training RF outperformed supervised RF only with ProtBert.

In contrast, self-training generally reduced SVM's ability to distinguish between antigenically similar and antigenically different virus pairs. Scores fell by 0.102--0.433 compared with supervised SVM. The sole exception was SVM using ProtVec with 25\% of training labels retained: self-training increased its score from 0.383 to 0.439.

Label spreading scored below supervised RF for every protein representation and label proportion (Figure~\ref{fig_results_all_groups}). Mean macro-F\textsubscript{1} for label spreading ranged from 0.768 to 0.788 with 25\% of labels retained and from 0.813 to 0.830 with all labels retained. ProtVec was the strongest representation for label spreading at every label proportion. With 25\% of training labels retained, label spreading with ProtVec achieved mean macro-F\textsubscript{1} of 0.788, close to the 0.791 achieved by supervised ProtVec-RF and the 0.798 achieved by self-training ProtVec-RF. With ProtVec, label spreading improved as the proportion of retained training labels increased from 25\% to 75\%, reaching 0.832. Retaining all training labels brought no further improvement (0.830).

Across all models, self-training ProtBert-RF achieved the highest mean macro-F\textsubscript{1} of 0.864 with all training labels retained, compared with 0.860 for supervised ProtBert-RF. With 75\% of training labels retained, self-training RF approached the classification performance of supervised RF trained with all available labels, although it remained slightly below it. The difference in mean macro-F\textsubscript{1} was 0.010--0.019 across the four representations.

\subsection{Comparison of Protein Representations}
\begin{figure}[!htb]
    \centering
    \includegraphics[width=0.8\linewidth]{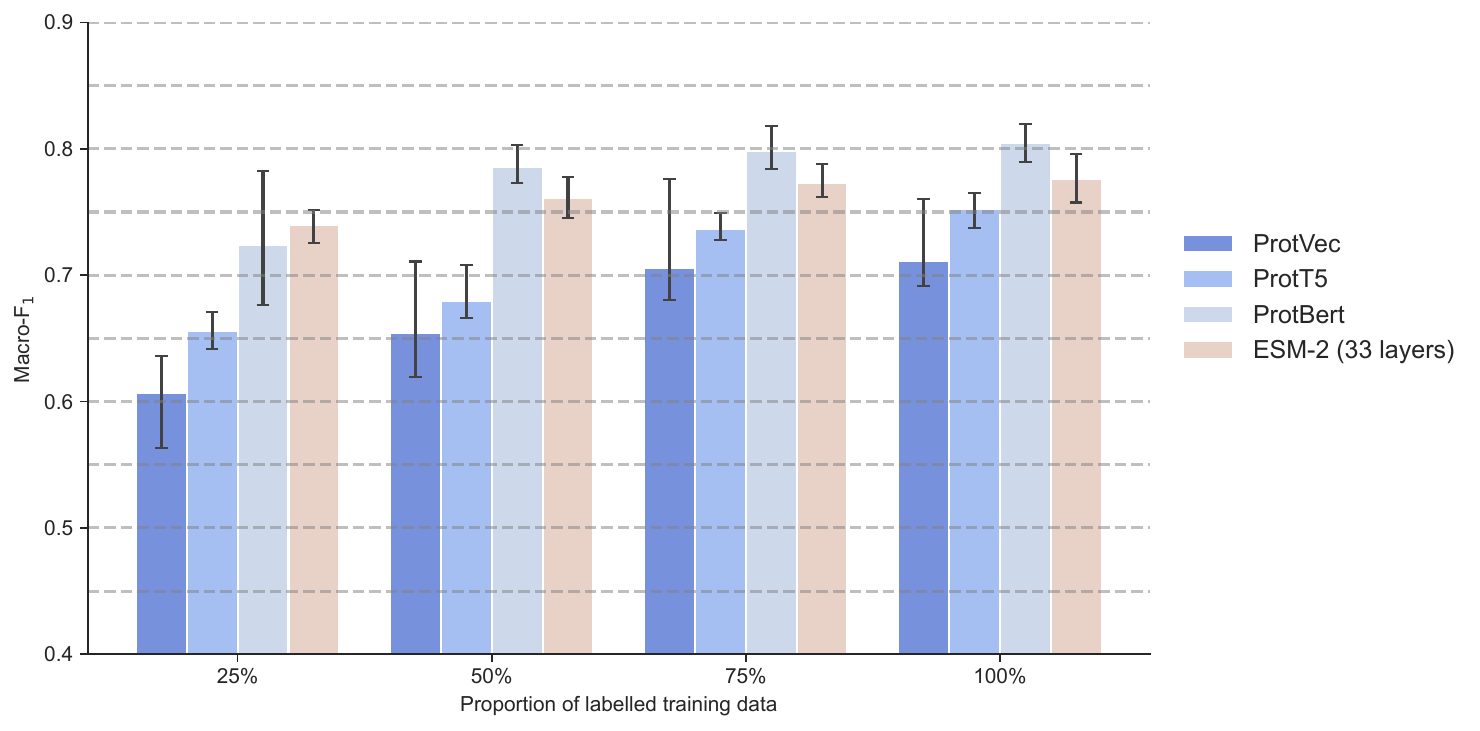}
    \caption{Mean macro-F\textsubscript{1} by protein representation and training-label proportion. Bars show the mean of the five fold-level model averages; error bars show their 2.5th--97.5th percentiles.}
    \label{fig_comp_pretrained}
\end{figure}

The best-performing protein representation changed with label availability (Figure~\ref{fig_comp_pretrained}). ESM-2 gave the strongest average classification performance with 25\% of training labels retained, scoring 0.740 compared with 0.724 for ProtBert, 0.656 for ProtT5, and 0.607 for ProtVec. ProtBert became the leading representation at higher label availability: it scored 0.786 with 50\% of training labels retained, 0.798 with 75\% retained, and 0.804 with all training labels retained. With all labels available, the corresponding scores for ESM-2, ProtT5, and ProtVec were 0.777, 0.752, and 0.711.

\begin{figure}[!htb]
    \centering
    \includegraphics[width=\linewidth]{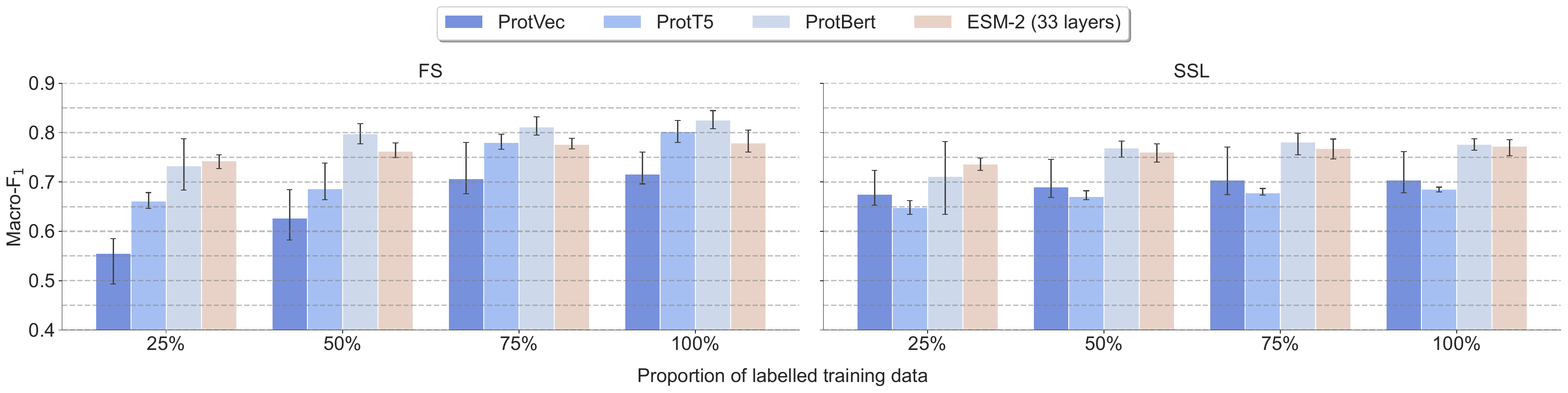}
    \caption{Mean macro-F\textsubscript{1} for supervised and semi-supervised models by protein representation and training-label proportion. Bars show the mean of the five fold-level group averages; error bars show their 2.5th--97.5th percentiles.}
    \label{fig_comp_pretrained_fs_ssl}
\end{figure}

On average, supervised models classified antigenic relationships better than SSL models when using ESM-2, ProtBert, or ProtT5 at every tested training-label proportion (Figure~\ref{fig_comp_pretrained_fs_ssl}). ProtVec was the exception when labels were scarce: SSL models performed better on average with 25\% or 50\% of training labels retained, but this advantage disappeared with 75\% or all training labels retained. With 25\% of training labels retained, the supervised and SSL averages for ProtVec were 0.556 and 0.675; with all labels retained, they were 0.716 and 0.704. For ProtT5, supervised models retained an advantage with all training labels available, scoring 0.802 on average compared with 0.685 for SSL models.

\begin{samepage}
ProtVec also performed well with RF, scoring higher than ESM-2 at 25\% label availability (Figure~\ref{fig_results_all_groups}). Supervised and self-training RF achieved 0.791 and 0.798 with ProtVec, compared with 0.776 and 0.778 with ESM-2. Its advantage extended to label spreading, for which it gave the highest score at every label proportion.

\end{samepage}

\Needspace{9\baselineskip}
\subsection{Performance across Subtypes}
\begin{figure*}[!htb]
\centering
\includegraphics[width=\linewidth]{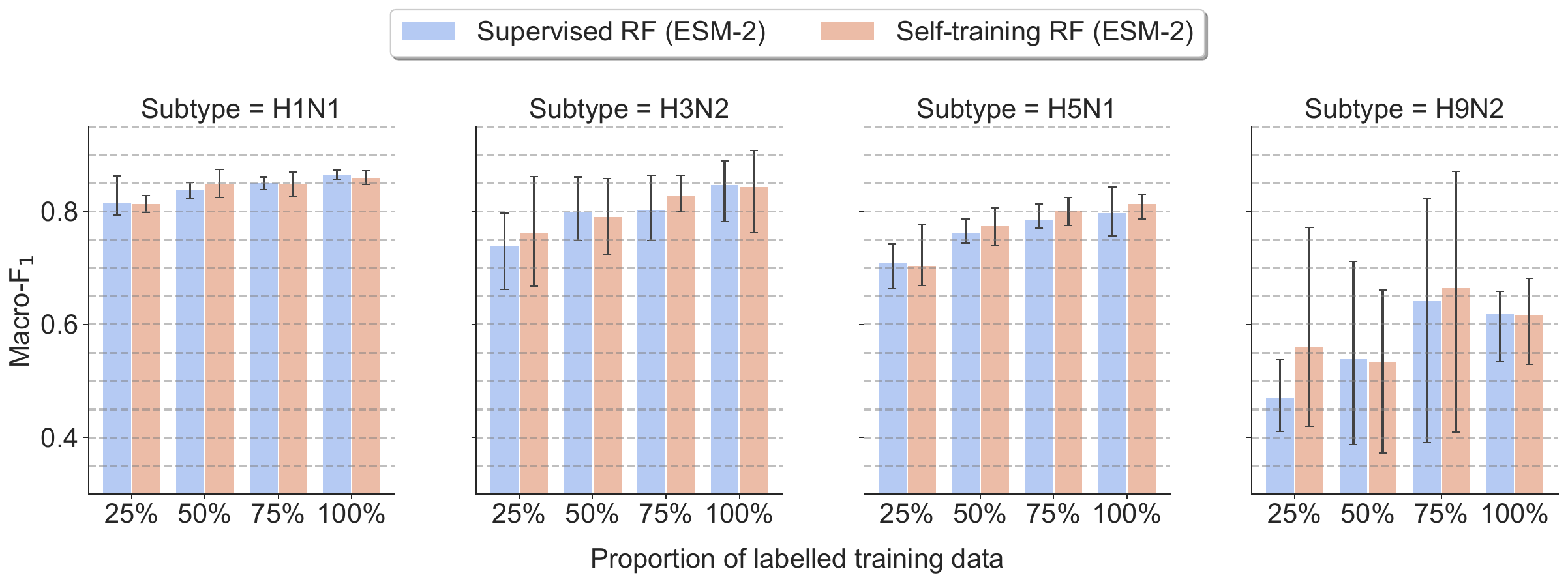}
\caption{Macro-F\textsubscript{1} for each subtype using supervised and self-training RF with ESM-2 embeddings. Bars show mean macro-F\textsubscript{1} across five outer test folds; error bars show the 2.5th--97.5th percentiles of the five fold scores.}
\label{fig_results_subtype_esm2_33}
\end{figure*}

\begin{figure*}[!htb]
\centering
\includegraphics[width=\linewidth]{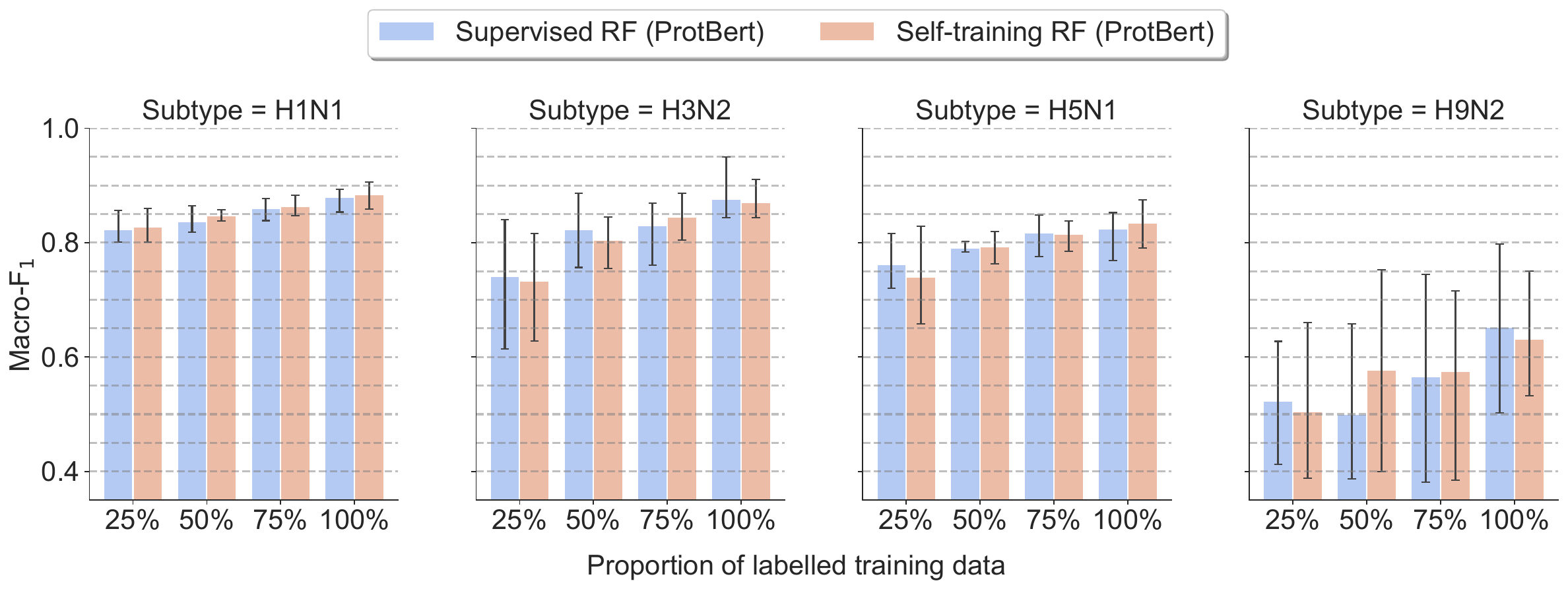}
\caption{Macro-F\textsubscript{1} for each subtype using supervised and self-training RF with ProtBert embeddings. Bars show mean macro-F\textsubscript{1} across five outer test folds; error bars show the 2.5th--97.5th percentiles of the five fold scores.}
\label{fig_results_subtype_protbert}
\end{figure*}

The RF models using ESM-2 or ProtBert performed differently across subtypes (Figures~\ref{fig_results_subtype_esm2_33} and \ref{fig_results_subtype_protbert}). Under both supervised learning and self-training, RF maintained relatively strong classification performance for H1N1 even with only 25\% of training labels retained, scoring 0.814--0.827. Using all training labels improved performance further, to 0.861--0.884.

\begin{samepage}
Additional labelled pairs improved supervised RF's ability to classify antigenic relationships in both H3N2 and H5N1. For H3N2, retaining all training labels rather than 25\% raised supervised RF scores from 0.739 to 0.848 with ESM-2 and from 0.741 to 0.876 with ProtBert. Across supervised learning and self-training, H3N2 scores ranged from 0.733 to 0.763 with 25\% of training labels retained and from 0.844 to 0.876 with all labels retained. When all training labels were available, supervised RF performed slightly better than self-training RF, which scored 0.844 with ESM-2 and 0.870 with ProtBert. H5N1 performance also improved when the proportion of retained training labels increased from 25\% to 100\%. Under supervised RF, scores increased from 0.709 to 0.799 with ESM-2 and from 0.762 to 0.824 with ProtBert. H5N1 scores across supervised learning and self-training ranged from 0.704 to 0.834.

\end{samepage}

H9N2 had the lowest mean macro-F\textsubscript{1}, ranging from 0.472 to 0.666. Self-training ESM-2 RF achieved the highest score with 75\% of labels retained. With 100\% of training labels retained, self-training gave no improvement over supervised RF: both scored 0.619 with ESM-2, while ProtBert-RF decreased from 0.652 to 0.632 under self-training.

\begin{samepage}
Most H9N2 errors involved misclassifying antigenically similar virus pairs as antigenically different. These RF models were less effective at recognising antigenically similar pairs than antigenically different pairs, with F\textsubscript{1} scores of 0.144--0.460 and 0.800--0.871, respectively. For ESM-2 RF with 25\% of training labels retained, self-training improved recognition of antigenically similar H9N2 pairs more than recognition of antigenically different pairs. Their respective F\textsubscript{1} scores increased from 0.144 to 0.307 and from 0.800 to 0.817. This raised macro-F\textsubscript{1} from 0.472 to 0.562, but antigenically similar pairs remained difficult to identify.

\end{samepage}

\FloatBarrier
\section{Conclusion and Discussion}

Supervised RF already classified antigenic relationships relatively well with fewer training labels. Self-training added only small gains, and those gains became less consistent when all observed training labels were retained. For SVM, self-training instead reduced the ability to classify antigenic relationships.

\begin{samepage}
ProtVec performed well with RF and outperformed the other representations when used with label spreading. Earlier stacking models used residue-level, epitope, and regional descriptors \citep{yin2018predicting}, while network embeddings combined sequence information with antigenic relationships \citep{peng2023prediction}. ProtVec \citep{asgari2015continuous} instead learns representations of short amino acid segments from a large protein collection. Although the contextual representations had higher overall averages, their advantage did not extend to every classifier. The choice of representation therefore needs to be considered together with the classifier used.

\end{samepage}

H1N1 retained relatively high RF scores when training labels were reduced, while H3N2 improved more clearly with additional labels and H9N2 remained difficult even with all labels available. The dataset contained many more labelled H1N1 pairs than H9N2 pairs, so predictions for the larger subtypes contributed more to the overall score. RF performed well overall, but its H9N2 scores remained low.

For H9N2, RF often misclassified antigenically similar virus pairs as variants. This pattern recurred with ESM-2 and ProtBert under both supervised learning and self-training, despite relatively high F\textsubscript{1} scores for antigenically different pairs. The small overall RF gains did not eliminate these errors in H9N2. H9N2 included 31 antigenically similar pairs and 87 antigenically different pairs, with only 3--9 similar pairs in each outer test fold. The small number of antigenically similar pairs in each test fold may partly explain the variation in H9N2 scores.



\section*{Acknowledgements}
This work was funded by the University of Liverpool.





\section*{Competing interests}
The author declares no competing interests.





\bibliographystyle{unsrtnat}
\bibliography{references}

\end{document}